\documentclass[letterpaper, 10 pt, conference]{ieeeconf}  % Comment this line out if you need a4paper

\IEEEoverridecommandlockouts                              % This command is only needed if 
\usepackage{amsmath}
\usepackage{amssymb}
\usepackage{pifont}% http://ctan.org/pkg/pifont

\usepackage{subcaption}
\usepackage{graphicx,adjustbox}
\usepackage{multicol}
\usepackage[font={small}]{caption}   %onehalfspacing
\usepackage{subcaption}
\usepackage{booktabs,makecell}
\usepackage{graphicx}
\usepackage{bm}
\usepackage{color}
\usepackage[dvipsnames]{xcolor}
\usepackage{float}
\usepackage{units}
\usepackage{soul}
\usepackage{comment}
\usepackage[breaklinks=true, bookmarks=true]{hyperref}
\usepackage{cite}
\usepackage{pifont}  % 放在导言区
\usepackage{caption}

\newcommand{\figref}[1]{Fig.~\ref{#1}}

\newcommand{\tabref}[1]{Table~\ref{#1}}
\usepackage{amsfonts}

\usepackage{xcolor}%
\usepackage{textcomp}%
\usepackage{manyfoot}%
\usepackage{booktabs}%
\usepackage{algorithm}%
\usepackage{algorithmicx}%
\usepackage{algpseudocode}%
\usepackage{listings}%
\usepackage{soul}
\usepackage{array}
\usepackage{comment}
\DeclareMathSizes{10}{10}{7}{5}  % Example for 10pt text size

\usepackage{booktabs,makecell}
\usepackage{supertabular} 
\usepackage{lipsum}       
\usepackage{longtable}
\usepackage{tabularx}
\usepackage{hyperref}
\usepackage{indentfirst}
\usepackage{ragged2e}
\usepackage{graphicx}%
\usepackage{multirow}%
\usepackage{amsmath,amssymb,amsfonts}%
\usepackage{amsthm}%

\DeclareMathSizes{10}{10}{7}{5}  % Example for 10pt text size

\usepackage{booktabs,makecell}
\usepackage{supertabular} 
\usepackage{lipsum}       
\usepackage{longtable}
\usepackage{tabularx}
\usepackage{hyperref}
\usepackage{indentfirst}
\usepackage{ragged2e}
\begin{document}
\title{\LARGE \bf
Hierarchical Intention-Aware Expressive Motion Generation for Humanoid Robots}

\author{Lingfan Bao, Yan Pan, Tianhu Peng, Dimitrios Kanoulas and Chengxu Zhou\textsuperscript{*}
\thanks{This work was partially supported by the Royal Society [grant number RG\textbackslash R2\textbackslash232409], the Advanced Research and Invention Agency [grant number SMRB-SE01-P06], and the UKRI FLF [MR/V025333/1] (RoboHike).  For the purpose of Open Access, the author has applied a CC BY public copyright license to any Author Accepted Manuscript version arising from this submission.}%
\thanks{The authors are with the Department of Computer Science, University College London, UK.}%
\thanks{\textsuperscript{*}Corresponding author, {\tt\small chengxu.zhou@ucl.ac.uk}}%
}%

\bstctlcite{IEEEexample:BSTcontrol}

\maketitle
\thispagestyle{empty}
\pagestyle{empty}
% Requirements
%The maximum number of pages per paper is 8 (including references)
%%%%%%%%%%%%%%%%%%%%%%%%%%%%%%%%%%%%%%%%%%%%%%%%%%%%%%%%%%%%%%%%%%%%%%%%%%%%%%%%
\begin{abstract}

Effective human-robot interaction requires robots to interpret human intentions and generate socially appropriate, expressive motions in real-time. However, few approaches effectively bridge the gap between nuanced social inference and the synthesis of physically embodied responses. To address this gap, we propose HIAER, a hierarchical framework that enables a robot to become intention-aware and respond with expressive motions. Our framework integrates an intention-aware vision language model using in-context learning to infer not only the primary social intent but also its affective context, providing Valence-Arousal estimates for adaptive decision-making. This inference then guides the generation of stylistically appropriate gestures in real-time. A reinforcement learning based whole-body controller ensures robust execution on a physical humanoid. In real-world experiments, our system produces behaviors that are rated as more socially intelligent and contextually appropriate, enabling more natural and effective human-robot interaction.

\end{abstract}

%%%%%%%%%%%%%%%%%%%%%%%%%%%%%%%%%%%%%%%%%%%%%%%%%%%%%%%%%%%%%%%%%%%%%%%%%%%%%%%%

\section{Introduction}
The cornerstone of effective human interaction is a tight coupling between inferring another’s intention and generating a corresponding, often non-verbal, response \cite{kendon2004gesture, cavallo2016decoding, 2023_nonverbal_hri}. This principle is also primary for humanoid robots operating in human-centric settings, where legible and adaptive gestures are not mere additions but essential components for building trust, facilitating collaboration, and enhancing user engagement \cite{2013_modeling_evaluation_humanoid_HRI,2024_humanoid_LLM_HRI_understandingLLMpowered}.

% \paragraph{The current gap:}
% HRI 中“意图→动作”链路缺少校准置信与歧义回退，且端到端延迟难以控制
% Expressive responses in social interaction can be analyzed through four components: a triggering event, its affective appraisal, an observable expression (e.g., face, voice, or posture), and action tendencies that guide behavior \cite{breazeal2003emotion}. 
Although recent advances have enabled humanoid robots to achieve robust locomotion \cite{2024_Bao_drl_bipedal_review} and to generate expressive motions \cite{2024_speratebody_expressive_allmotion,he2025asap}, these capabilities are rarely grounded in the immediate social context or calibrated through confidence-aware selection. As a result, even physically sophisticated movements often lack sensitivity to the unfolding interaction, causing robots to move without conveying meaning and to miss the timely, legible, and adaptive non-verbal behaviors that are essential for effective human–robot interaction.

Motivated by these limitations, we present HIAER: Hierarchical Intention-Aware Expressive Responses for Humanoid Robots, a framework that generates socially appropriate gestures by hierarchically interpreting both the social intent and affective context of human behavior. At its core, HIAER first uses a vision-language model (VLM) with in-context learning (ICL) \cite{2022_incontextlearning_rethinkingroleofdemonstration,2024_ICL_EMOTION_humanoid_VLM_demonstration} to infer the social intent of the human, establishing the functional goal of the interaction. As an integral part of its reasoning process, the VLM is prompted to also assess the scene's affective context in terms of valence–arousal (V–A) \cite{vamodel1,vamodel2}, using these dimensions to modulate the expressive style of the final motion response. Conditioned on prompts derived from this hierarchy, a text-to-motion diffusion model \cite{2025_MDM_DART} synthesizes socially appropriate gestural motions in real time, which are executed on a physical robot via a low-level reinforcement learning (RL)-based whole-body controller.

% \paragraph{Paper Focus and framework explaination}
% Motivated by the limitations, we present HIAER: Hierarchical Intention-Aware Expressive Responses for Humanoid Robots,a framework that couples calibrated intention inference with a valence–arousal (V–A) affect estimator to deliver socially appropriate gestural responses while ensuring polite, safe, and predictable behavior under ambiguity. HIAER integrates a vision-language model (VLM) with in-context learning  to infer human intent from prompts and body language, and to evaluate the recognized interaction scene using calibrated confidence and V–A estimates \cite{breazeal2003emotion}, thereby selecting and parameterizing the final gestural response. Conditioned on motion prompts, a text-to-motion diffusion model \cite{2025_MDM_DART} synthesizes temporally coherent, socially appropriate motion sequences in real time, which are executed on physical hardware via a low-level whole-body controller.

% Our system empowers humanoid robots to produce socially aligned, physically executable gestures in real time. 
The key contributions of this work are: 

\begin{itemize}
    \item  We propose a hierarchical framework that integrates a VLM's fine-grained reasoning on both social intent and affective context to generate socially aware motions on a humanoid robot.

    % \item We architect a VLM-with-ICL intention module that infer intent, context, and calibrated confidence;  conditioned on this inference, we derive V-A estimates that parameterize the selected motion primitive and its expressivity, while confidence gates selection and fallbacks under ambiguity.

    \item We introduce estimation of the V-A model to parameterize complex social intentions into actionable parameters, which in turn guide the selection of context-appropriate expressive motions.

    \item We design a VLM-based social intention-aware module that leverages ICL and Chain-of-Thought (CoT) \cite{CoT} prompting to derive V-A estimates, which in turn guide the selection of a corresponding motion.

    \item The implementation and validation of the complete system on a physical humanoid robot in real-world HRI scenarios, demonstrating its capability to produce low-latency, context-appropriate gestures.

\end{itemize}

The structure of this paper is as follows.  Section~\ref{sec:related_work} reviews related work.  Section~\ref{sec:framework_overall} presents the HIAER framework, including intention inference and V–A conditioning motion selection, and whole-body motion execution. Section~\ref{sec:experiments} describes the experimental setup and reports quantitative and qualitative results in representative HRI scenarios, and discusses the limitation. Section~\ref{sec:conclusion} concludes and outlines future work.

\begin{figure*}[t]{
      \centering
      % \vspace{.06in}
      % \vspace{-1em}
      \includegraphics[trim = 0mm 105mm 90mm 0mm, clip, width=\linewidth]{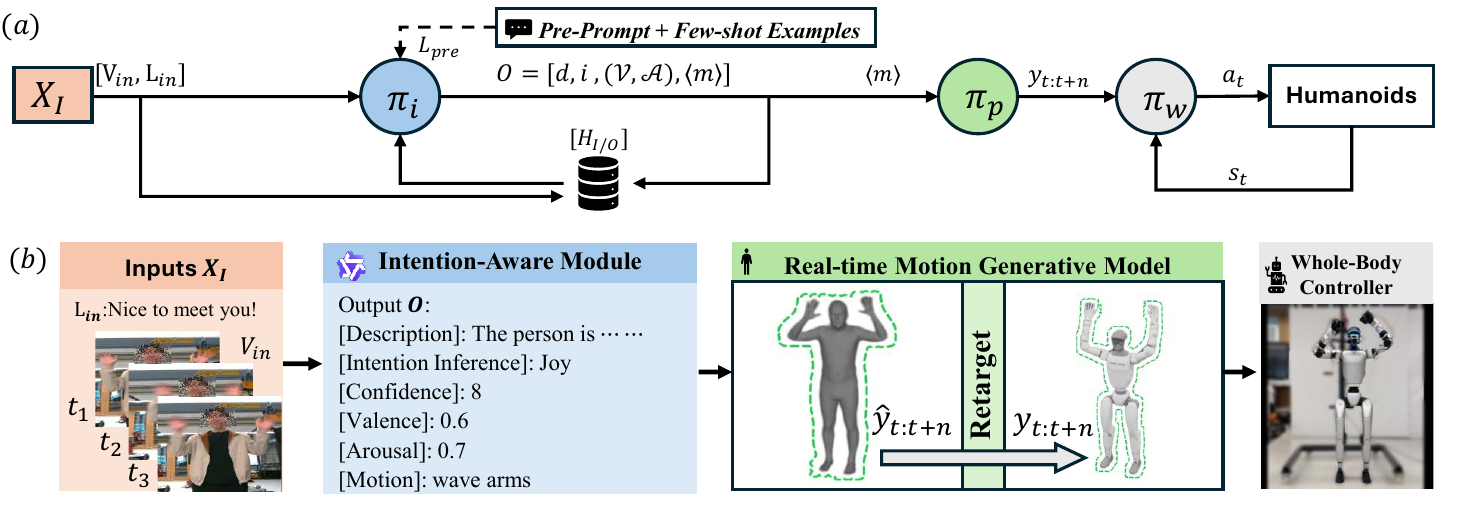} \\
      % \vspace{-0.5em}
      \caption{\textbf{Overall framework of the proposed work.} (a) The high-level system architecture. Multimodal inputs $X_I=(V_{in},L_{in})$ are parsed by the intention-aware module $\pi_i$ (with pre-prompt, few-shot, and I/O history) to produce a structured output $O$ and motion primitive $\langle m\rangle$, which condition the planner $\pi_p$ and whole-body controller $\pi_w$ to generate actions $a_t$. (b) A concrete example of the pipeline in an interaction scenario: $\pi_i$ outputs description, intent, confidence, and V–A; the motion generator synthesizes and retargets the gesture for execution on the humanoid.
      \label{fig:overall_framework}
      }}
      % \vspace{-0.2cm} 
\end{figure*}

% The system is designed to interpret multi-modal human cues and generate socially appropriate, real-time motion responses on a humanoid robot. (a) The high-level system architecture. Multi-modal inputs $X_{I}$, consisting of vision $V_{in}$ and prompt $L_{in}$, are processed by the Intention-Aware Module $\pi_i$. Guided by a pre-prompt, few-shot examples, and its I/O history $H_{I/O}$, this module produces a structured output $O$ containing the inferred intent within inner CoT chain, and selected motion $\langle m \rangle$}. This primitive guides a motion planner $\pi_p$ and a whole-body controller $\pi_w$ to generate the final robot action $a_t$. Dashed arrows denote one-time initial inputs, while solid arrows indicate continuous data flow. (b) A concrete example of the pipeline in an interaction scenario. The framework takes raw inputs, and the Intention-Aware Module (blue box) generates a detailed Chain-of-Thought output, including confidence and V-A scores and selected prompt is then synthesized by the Real-time Motion Generative Model (green box), retargeted, and executed on the physical robot.

\section{Related Work}
\label{sec:related_work}

\subsection{Expressive Human-Robot Interactions}

Non-verbal communication is a cornerstone of human interaction, conveying complex emotions, intentions, and social cues that transcend spoken language \cite{kendon2004gesture,cavallo2016decoding,2023_nonverbal_hri}. Enabling robots with such expressive capabilities is critical for effective human-robot interaction, as it significantly enhances communication quality and user engagement. Traditional approaches to generating expressive behaviors have relied on rule-based \cite{2020_rule_based_Cassie_manual_interface,2013_HRI_rulebasedmethod_behaviorgeneneration,2019_HRI_rulebased_Bodystorming} or template-based methods \cite{template1}. While predictable, these methods are inherently rigid and struggle to adapt to the fluid, dynamic nature of real-world social contexts \cite{2024_ICL_EMOTION_humanoid_VLM_demonstration}.

The recent and rapid advancement of VLMs has unlocked new potential for robots to understand and reason about high-level human intentions with unprecedented sophistication \cite{VLM_intention1,VLM_intention2}. These models have shown great promise in identifying human goals and decomposing them into actionable tasks, particularly in fields like legged robotics for locomotion planning \cite{2024_LLM_Prompt2Walk,2024_LLM_humanoidrobot_ComprehensiveLocomotionControl}. However, a critical gap remains: their focus has largely been on interpreting explicit, functional goals rather than the implicit, affective intentions embedded within a dynamic social environment. Consequently, their output is often limited to task decomposition, failing to complete the intention-to-motion loop with a response that is not just functional, but physically and socially expressive.

In contrast, our framework, HIAER, is specifically designed to provide a complete closed-loop interaction by integrating rich affective reasoning with embodied execution. Our system is able to not only understand a user's functional intent, but also to assess the scene's affective context through VA estimates.

\subsection{Expressive Motion Generation}
% 1. motion generation 很重要 2, 现在生成主要有两种方法 - retrieval and generative model 的方式 3. retrieval ... 然后 generative model 怎么样好处什么 4. 现在的数据集能够支持小模型训练动作十分完备，而且 t2m 形式的训练以及数据 提供了很好的 界面或者接口 5. 现在的框架 EMOTION and GenEM 并没有很好的利用到这些数据，相反还要通过demonstrate 或者predefine的方法 会导致动作并没有那么连贯以及自然 6. 我们的框架 利用了 DART ..

Generating expressive and natural motions is critical for effective human-robot interaction, as it enables robots to clearly convey social intent and respond dynamically in real-world contexts. Current data-driven approaches broadly include retrieval-based methods and generative models. Retrieval methods select motion sequences from existing databases, offering realism and quick inference but limited flexibility due to library constraints.

Generative models, such as GANs \cite{2018_T2A_GANs}, VAEs \cite{2020_action2motion_VAEs}, and diffusion models \cite{2020_diffusion_model}, allow for novel and contextually adaptable motions but often require intensive training and computational resources. Diffusion-based models, in particular, have demonstrated exceptional capabilities in generating diverse, high-quality motions, such as MDM \cite{2023_tevet_MDM}. Recent models like DART \cite{2025_MDM_DART} further support real-time performance through efficient latent-space denoising conditioned on textual descriptions and spatial constraints.

% Large-scale motion datasets, such as AMASS \cite{2019_Naureen_AMASS}, HumanML3D \cite{2022_Dataset_HumanML3D_Guo_CVPR} facilitate training compact yet powerful generative models, effectively bridging the gap between high-level human intent and low-level motion execution. However, existing frameworks like EMOTION++ \cite{2024_ICL_EMOTION_humanoid_VLM_demonstration} and GenEM \cite{2024_GenerativeExpressiveBehavior_LLM} still rely significantly on demonstration-based methods or predefined templates, limiting their adaptability and naturalness.

The availability of large-scale motion datasets, such as AMASS~\cite{2019_Naureen_AMASS} and HumanML3D~\cite{2022_Dataset_HumanML3D_Guo_CVPR}, has been pivotal in training powerful, generalized generative models. These extensive archives provide a rich prior of human movement, enabling models to synthesize a vast array of motions from high-level commands, thereby avoiding the need for time-consuming imitation learning for each new desired action. However, despite this progress, existing frameworks like EMOTION++ \cite{2024_ICL_EMOTION_humanoid_VLM_demonstration} and GenEM \cite{2024_GenerativeExpressiveBehavior_LLM} still rely significantly on specific demonstration-based methods or predefined templates. This dependency ultimately limits their adaptability and naturalness, as they are constrained by the scope of the provided examples rather than a broader, learned understanding of human motion.

% Our proposed approach leverage a diffusion-based generative model DART \cite{2025_MDM_DART} with RL-based WBC, training on these retargeted datasets. By capitalizing on structured datasets and a streamlined text-to-motion interface, our framework produces diverse, socially appropriate, and physically executable motions, enhancing real-time humanoid robot interactions.

Our proposed approach is built on a pipeline that pairs the diffusion-based generative model DART \cite{2025_MDM_DART}  with a RL-based WBC. Crucially, both components are grounded in large-scale human motion datasets. This deep foundation in real human data is what empowers our framework to generate a diverse and expressive repertoire of motions from high-level commands, without relying on limited, predefined templates.

\section{Hierarchical Intention-aware Expressive Framework}
\label{sec:framework_overall}
This section details our hierarchical framework, HIAER, designed to couple high-level intention inference with low-level expressive motion execution part, thereby enabling interactions that are responsive to social context.

\subsection{Framework Overview}

As illustrated in Fig.~\ref{fig:overall_framework}, our framework is centered around the Intention Awareness Module, denoted $\pi_i$. This module is implemented as a VLM agent that processes multi-modal inputs $X_I = (V_{in},L_{in})$ along with conversational history $H_{I/O}$. Guided by an ICL-style instruction $L_{pre}$ and a CoT prompting strategy, the module performs a holistic analysis of the interaction. This single reasoning process culminates in a structured output $O$, as
\begin{equation}
O = [d, i, c, (\mathcal{V},\mathcal{A}), \langle m \rangle] = \pi_i(L_{pre}, X_I, H_{I/O}), 
\end{equation}
which contains a scene description $d$, the inferred human intent $i$ with a confidence score $c$, a V–A estimate $(\mathcal{V},\mathcal{A})$ of the affective context, and a corresponding motion primitive $\langle m \rangle$.

The high-level command $O$ generated by $\pi_i$ guides the mid-level motion planner $\pi_p$. This planner, implemented as a text-to-motion diffusion model \cite{2025_MDM_DART}, synthesizes a temporally coherent trajectory of human motion trajectories, $\hat{y}_{t:t+n}$, where $n$ is timestep of future prediction horizon. Subsequently, this trajectory is retargeted to the robot's specific kinematics, resulting in a desired trajectory $y_{t:t+n}$. Finally, this trajectory is passed to the low-level whole-body controller, $\pi_w$, which acts as a reference tracking and balancing controller to generate executable joint commands $a_t$,  incorporating state feedback to ensure $s_t$ stable and safe execution.

Through this hierarchical design, the framework bridges high-level social reasoning with low-level, dynamic whole-body control. This integration enables our humanoid robot to produce expressive and socially appropriate responses in real time, a critical capability for effective human-robot interaction.

% The system takes as input a user image observation $O_{\text{in}}$ and a natural language prompt $l_{\text{in}}$. These inputs are first processed by the \textit{Intention-Aware Module}, which serves as the strategy-level component of the framework. This module outputs a structured motion prompt, decomposed into a sequence of short expressive motion clips $m_i$, each representing a temporally coherent gesture segment. Additionally, both input and output information are logged into a rolling interaction history buffer denoted as $H_{\text{I/O}}$, which supports iterative refinement and context-aware reasoning across time.

% These motion clips are passed into the real-time generation model DART, which synthesizes human skeleton motion sequences as reference trajectories. These reference motions are then retargeted to the G1 humanoid robot model. To ensure smooth and responsive execution, we perform high-frequency interpolation on the $25$ FPS motion output, upsampling it to $500$ Hz. The resulting joint trajectories are temporally aligned and converted into continuous position references. The DART module operates online, continuously generating a short motion horizon $\hat{x}_{T:T+8}$, which is fed into a low-level PD tracking controller to compute real-time torque commands $\tau_t$.

\begin{figure}{
      \centering
      % \vspace{.06in}
      % \vspace{-1em}
      \includegraphics[trim = 3cm 0cm 5cm 0cm, clip, width=\linewidth]{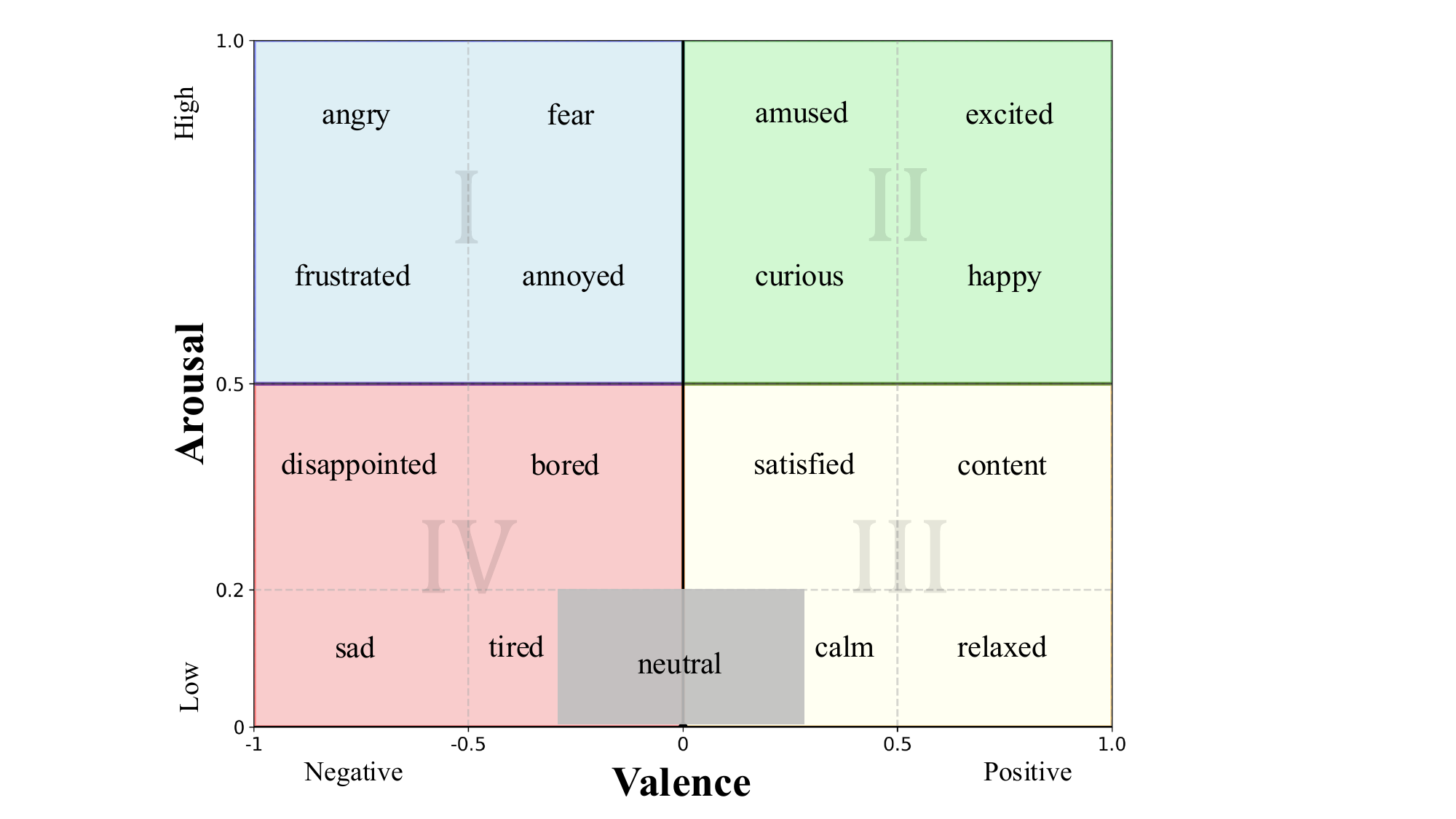} \\
      % \vspace{-0.5em}
      \caption{Our framework leverages V-A space to interpret human affective context and modulate the robot's response. The blue area (Quadrant I) represents states of high Arousal and negative Valence, while the green area (Quadrant II) corresponds to high Arousal and positive Valence. Conversely, the yellow area (Quadrant III) signifies low Arousal and positive Valence, and the red area (Quadrant IV) indicates low Arousal and negative Valence. The grey represents the Neutral state of non-activation.}
      \label{fig:VA_demo}
      }
      \vspace{-0.4cm} 
\end{figure}

\subsection{Intention-Aware and Inference Module}
\label{subsec:intention_aware_module}

This section details the design of our core perception and reasoning module, $\pi_i$, which interprets human behavior from multi-modal inputs. The module is built upon the pre-trained Qwen2.5-VL-32B \cite{Qwen2.5_VL}. The VLM is guided by a meticulously designed ICL strategy that programs its behavior at runtime via a carefully constructed prompt, $L_{pre}$.

\paragraph{V-A model for Expressiveness} A key component of the module's output is the estimation of the interaction's affective context, represented using the V-A model. The V-A model is a widely adopted psychological framework for representing emotional states along two continuous axes: Valence, representing the pleasure-displeasure continuum, and Arousal, representing the activation-deactivation continuum, as illustrated in Fig.~\ref{fig:VA_demo}. By prompting the VLM to map observable cues $(d, i, c)$  to this two-dimensional space, we can effectively modulate the expressive style of a selected motion primitive $(\mathcal{V},\mathcal{A})$, which can show expressiveness of emotion. For instance, the primitive wave right arm, when coupled with a high-Valence, medium-Arousal state, results in a friendly and welcoming gesture, rather than a robotic or lifeless one.

\paragraph{Pre-prompt Design}: To generate this nuanced and structured output, as one of the important components of $L_{pre}$, pre-prompt is structured around four key components, and enforced with CoT. 

To generate this nuanced and structured output, the pre-prompt within $L_{pre}$ defines the VLM's core instructions, constraints, and the required reasoning structure. It is composed of three main parts:

\begin{itemize}
    \item Persona and Task Definition: The prompt first establishes the robot's persona as a helpful assistant and defines its overarching task: to perform a multi-step analysis of the human-robot interaction.
    \item Output Format and CoT Enforcement: It then enforces a rigid output specification, requiring a structured format with six key fields: Description $d$, Intention Inference $i$, Confidence $c$, V-A values $(\mathcal{V},\mathcal{A})$, and Motion Primitive $\langle m \rangle$, as shown in Fig.\ref{fig:overall_framework} (b). Critically, this part also explicitly instructs the model to follow the CoT reasoning path to generate these fields sequentially.
    \item V-A Context mapping. To anchor the numerical estimation of the affective state, the prompt includes a mapping table that provides a rich semantic context. This table grounds each quadrant of the V-A space (shown in Fig.~\ref{fig:VA_demo}) by associating it with specific target ranges for Valence and Arousal, as well as a set of characteristic physical gestures. For instance, Quadrant I is linked to gestures of aggression and tension; Quadrant II to welcoming gestures; Quadrant III to gestures of calm and support; and Quadrant IV to defensive or withdrawal gestures. By providing these explicit guidelines that connect abstract affective states to both numerical ranges and observable actions, we constrain the model's output, ensuring that the generated $(\mathcal{V},\mathcal{A})$ values are more consistent and meaningfully grounded in the social context.
    
    % V-A Context mapping: To anchor the numerical estimation of the affective state, the prompt includes a mapping table that provides semantic context. i) Quadrant I is identified using gestures that indicate aggression or tension, ii) Quadrant II is identified using gestures that indicate an open and welcoming attitude, iii) Quadrant III is identified using gestures that indicate supportiveness and calm, and iv)  is identified using gestures that indicate defensiveness or withdrawal. with specific target ranges for Valence and Arousal shon in the Fig~\ref{fig:VA_demo}. By providing these explicit guidelines, we constrain the model's output, ensuring that the generated $(\mathcal{V},\mathcal{A})$ values are more consistent and meaningfully grounded in the social context.

    \item Safety Rules: Finally, the pre-prompt includes a set of safety rules, chief among them a fallback mechanism that defaults to a neutral gesture when inference confidence is low, and a strict prohibition against aggressive or unsafe motions.
\end{itemize}

% First, it established the robot's persona as a helpful assistant and defines its overarching task: to perform a multi-step analysis of the interaction. Second, it enforces a rigid output format specification: The output of the inference module, $\pi_i$ is enforced to be in a structured format containing six key fields: Description $d$, Intention Inference $i$, Confidence $c$, V-A values $(\mathcal{V},\mathcal{A})$, and Motion Primitive $\langle m \rangle$, which is illustrate in the Fig.\ref{fig:overall_framework} (b), to ensure reliable parsing by downstream modules. Besides, in this part, the CoT is enforced there.

% Third, it includes a set of safety rules, chief among them a fallback mechanism that defaults to a neutral gesture when inference confidence is low, and a strict prohibition against aggressive motions. 

\paragraph{Few-shot Examples} 
Complementing the instructions in the pre-prompt, the few-shot examples are a critical part of $L_{pre}$ that provide concrete demonstrations of the desired behavior. Each example presents a full scenario, mapping multi-modal inputs (e.g., an image and a text prompt) to a complete, structured output that follows the CoT logic. These examples are vital for grounding the abstract tasks. For each demonstration, the $(\mathcal{V},\mathcal{A})$ values and corresponding motion primitive $\langle m \rangle$ were manually annotated by the authors based on our interpretation of the situational description and intent. By providing a clear baseline of these expert-labeled examples, the VLM learns to reliably adhere to the intended reasoning path and output format.

\begin{figure}[t]{
      \centering
      % \vspace{.06in}
      % \vspace{-1em}
      \includegraphics[trim = 0cm 2.2cm 0cm 1.4cm, clip, width=0.6\linewidth]{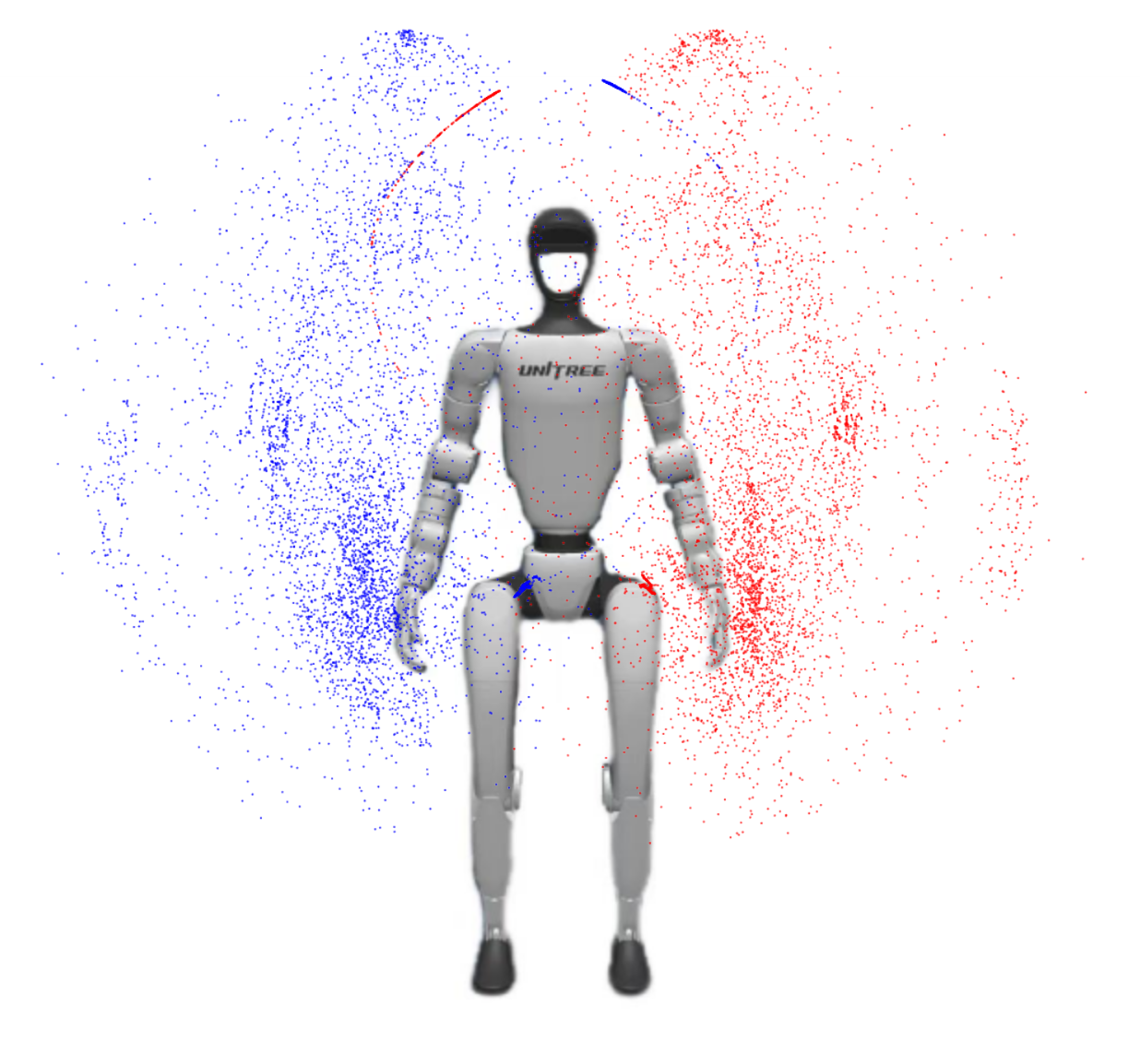} \\
      % \vspace{-0.5em}
      \caption{Balanced wrist workspace distribution for training.}
      \label{fig:motion distribution}
      }
      \vspace{-0.4cm} 
\end{figure}

\subsection{Expressive Motion Generation}
%在系统中，从语言prompt(结构动作片段)到动作生成模型输出动作参考轨迹的全过程。
% \begin{item}
%     \item Intro， DART is SOTA model. Why we choose DART?
%     \item Input/Output
%     \item training data
%     \item Inference Design
%     \item Motion retargeting
% \end{item}

The mid-level motion planner,$\pi_p$, serves as a flexible interface between the high-level intention-aware from $\pi_i$, and humanoid robot motion execution, translating the high-level textual motion command $\langle m \rangle$ into a kinematically feasible trajectory $\hat{y}_{t:t+n}$ for the humanoid robot.

This translation is achieved via two-stage process: (i) generating a realistic human motion sequence from the text prompt, and (ii) retargeting this human motion onto the robot's morphology.

% serves as a flexible interface between the high-level intention-aware from $\pi_i$, and humanoid robot motion execution. Its core function is to translate abstract commands into feasible, expressive, and temporally coherent motion trajectories.  

\paragraph{Human Motion Generation}
For the human motion generation stage, we employ the DART model~\cite{2025_MDM_DART}, a text-to-motion diffusion model. It was trained on large-scale human motion datasets, including HumanML3D~\cite{2022_Dataset_HumanML3D_Guo_CVPR} and BABEL~\cite{BABEL}, which source their motion data from the comprehensive AMASS archive~\cite{2019_Naureen_AMASS}. The framework operates auto-regressively to ensure temporally coherent motion. At each step, the model is conditioned on both the textual motion primitive $\langle m \rangle$ (e.g. ``wave arms''), and the preceding motion clip $\hat{y}_{t-n:t-1}$ to synthesize the next motion sequence, $\hat{y}_{t:t+n}$, as
\begin{equation}
\label{eq:dart}
 y_{t:t+n} = \pi_i(\langle m \rangle,\hat{y}_{t-n:t-1}).
\end{equation}

To ensure the generated motions are stable and executable by the robot, we implement the module with specific operational constraints. It runs online using a moving-window approach at a frame rate of 12.5 FPS, a choice that promotes smoother dynamics and aids in maintaining the robot's balance. Furthermore, to guarantee a stable start, the system is initialized with a 4-frame ``stand'' pose before the first commanded action is rolled out. The predicted window $n$ is setting as $8$ frames. This setup enables the real-time generation of natural and seamless human-like motions suitable for interactive scenarios.

\paragraph{Motion Representation and Retargeting} Our framework first synthesizes a human motion sequence $\hat{y}_{t:t+n}$, which serves as the source representation for retargeting. This sequence is based on the SMPL model \cite{SMPL} using a $22$-joint skeleton. Each frame is described by a $135$-dimensional vector, where ${q}_{\text{SMPL}} \in \mathbb{R}^{135}$ consists of a root translation $r \in \mathbb{R}^3$ and a continuous 6D \cite{2019_Cont6d} rotation representation for 22 joints, i.e., $q_{cont6d} \in \mathbb{R}^{22 \times 6}$. The target representation is the desired trajectory $y_{t:t+n}$, for our hardware platform, the 29-DoF Unitree G1 humanoid robot~\cite{unitree_g1}.

% The human motion sequence $(\hat{y}_{t:t+n})$, synthesized by the DART module, serves as the source for retargeting. This sequence is based on the SMPL model using a $22$ body joints skeleton with $1$ root. For each frame, the human pose is described by a 135-dimensional vector, $q_{\text{SMPL}}\in \mathbb{R}^135$. This vector comprises a 3D root translation and a continuous 6D rotation representation for each of the 22 joints.

% The target platform for our framework is the Unitree G1 humanoid robot~\cite{unitree_g1}. It stands 130 cm tall and features 29 degrees of freedom (DoF). The goal of retargeting is to a target state for the robot's 29 articulated joints. 

To bridge the morphological gap in real-time, we employ a neural retargeting network $\mathcal{X}$, implemented as a MLP with two hidden layers of $512$ units, trained via supervised learning. The network performs a frame-wise mapping from the human pose to the robot's full kinematic state. For each frame, it takes the human pose vector $q_{\text{SMPL}}$ as input, learns to distill the core motion intent while discarding human-specific articulations, and outputs the target state for all $29$ of the robot's actuated joints.

\begin{table}[t]
\centering
\scriptsize
\setlength{\tabcolsep}{2pt}
\caption{Training randomization parameters and reward function components.}
\label{tab:training_config}
\begin{tabular}{@{}c@{\hspace{0.5cm}}c@{}}
\begin{minipage}[t]{0.45\linewidth}
\centering
\vspace{0pt}
\begin{tabular}{lc}
\toprule
\textbf{Randomization} & \textbf{Range} \\
\midrule
Reference state init & Motion frames \\
External force & $\pm 3$ N \\
External torque & $\pm 0.5$ Nm \\
Friction coeff. & $[0.3, 1.0]$ \\
Base mass & $[-1, 3]$ kg \\
Angular velocity & $\pm 0.2$ rad/s \\
Joint position & $\pm 0.01$ rad \\
Joint velocity & $\pm 1.5$ rad/s \\
\bottomrule
\end{tabular}
\end{minipage}
&
\begin{minipage}[t]{0.45\linewidth}
\centering
\vspace{0pt}
\begin{tabular}{lc}
\toprule
\textbf{Reward Term} & \textbf{Weight} \\
\midrule
Joint pos tracking & 1.25 \\
Alive bonus & 0.25 \\
Action rate & -0.05 \\
Joint limits & -5.0 \\
Orientation & -5.0 \\
Base height & -10.0 \\
Feet sliding & -0.2 \\
Undesired contacts & -1.0 \\
\bottomrule
\end{tabular}
\end{minipage}
\end{tabular}
\end{table}

\begin{figure*}[t]{
      \centering
      % \vspace{.06in}
      % \vspace{-1em}
      \includegraphics[trim = 0mm 60mm 0mm 0mm, clip, width=\linewidth]{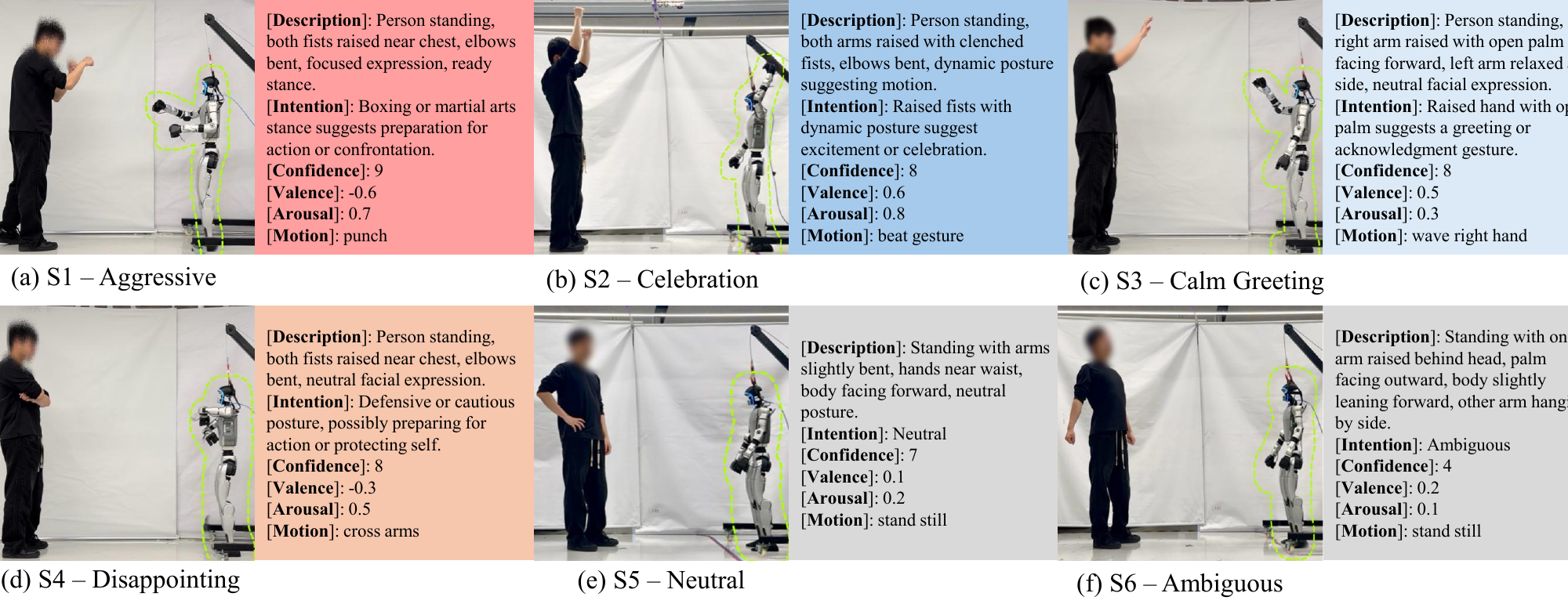} \\
      % \vspace{-0.5em}
      \caption{\textbf{Qualitative results across the six representative interaction scenarios.}  Each subfigure from (a) to (f) displays the human's input gesture (left), the robot's physical response (right), and the intermediate output generated by the Intention-Aware Module $\pi_i$
      . The output box details the system's internal reasoning, including the scene description, inferred intent, confidence score, Valence-Arousal (V-A) estimates, and the selected motion primitive. Then the selected motion primitive that is then executed by the humanoid robot.
      }\label{fig:HRI_Scenario}}
      % \vspace{-0.2cm} 
\end{figure*}

\subsection{Whole-Body Controller}
This section details our whole-body control policy $\pi_w$ design, which tracks reference motions from the generative policy $\pi_p$ while maintaining robot balance. We implement this controller using RL to achieve robust expressive motion execution.

The trained policy $\pi_{w}: \mathcal{O} \rightarrow \mathcal{A}$ maps observations $o_t = [s_t^{\text{root}}, q_t, \dot{q}_t, a_{t-1}, y_t]$ to target joint angles $a_t \in \mathbb{R}^{29}$, where $s_t^{\text{root}}$ denotes root state, $(q_t, \dot{q}_t)$ are joint positions and velocities, $a_{t-1}$ is the previous action, and $y_t$ is the generated reference from $\pi_p$. A PD controller then actuates the motors to track these targets $a_t$.

The training motion dataset is constructed by optimization-based retargeting of AMASS motions into Unitree G1 compatible trajectories \cite{he2024learning}. To prevent the policy from overfitting to frequently occurring poses, we balance resample the training dataset to achieve uniform wrist workspace coverage. Additionally, we incorporate all trajectories generated by $\pi_p$ into the training dataset to provide rich motion trajectory materials. As shown in Fig.~\ref{fig:motion distribution}, this ensures diverse wrist positions during training, improving generalization.

Policy training is performed in Isaac Lab \cite{isaaclab}, utilizing GPU-parallelized simulation with 4096 environments. The training converges after approximately 68,000 epochs, requiring around 14 hours on a single NVIDIA GeForce RTX 5090 GPU.

We employ curriculum learning where episodes terminate when joint tracking error $\|q_t - q_t^{\text{ref}}\| > \epsilon_{\text{term}}$, with $\epsilon_{\text{term}}$ progressively increasing throughout training to enforce more precise motion tracking. Training incorporates domain randomization and a multi-objective reward function detailed in Table~\ref{tab:training_config}.

\section{Experiments}
\label{sec:experiments}
We conduct a comprehensive set of experiments to validate the performance and capabilities of our HIAER framework. Our evaluation is designed to answer the following questions:

\begin{itemize}
    \item Q1: How accurately does HIAER’s hierarchical reasoning module infer human intent and affective context, and subsequently select a contextually appropriate motion?
    \item Q2: Does HIAER generate responses that human evaluators perceive as socially appropriate and expressive?
    \item Q3: Is HIAER sufficient for fluid real-time interaction, and is the system robust enough to operate reliably in dynamic environments with unscripted events?
\end{itemize}

Our evaluation is structured in three stages: first, we quantitatively assess the framework's core performance and internal validity; second, we perform ablation studies to justify key design choices; and finally, we test the system's robustness in challenging, real-world scenarios to demonstrate its generalization capabilities.

\subsection{Experimental Setup}
\label{subsec:setup}
In this section, we details our experiment setup. We firstly introduce our hardware, then the six interaction scenarios have been proposed to validate our framework.

\paragraph{Hardware} All experiments were conducted using the Unitree G1 robot \cite{unitree_g1}, equipped with an Intel RealSense D405 RGB-D camera on its head for visual perception. The core reasoning and motion generation modules were executed on a ground-station workstation with an AMD Ryzen 9 9950X CPU and an NVIDIA GeForce RTX 5090 GPU. Our Intention-Aware Module $\pi_i$ is built upon the pre-trained Qwen2.5vl:32B model \cite{Qwen2.5_VL}, while the motion planner $\pi_p$ utilizes the pre-trained DART diffusion model. Both models are deployed locally on our experimental workstation, ensuring real-time performance and independence from external cloud services.

\paragraph{Interaction Scenarios}
While daily interactions are diverse, they can be categorized by their non-verbal, emotional expression \cite{daily_emotion}. We therefore designed our evaluation around a set of six targeted scenarios. To evaluate expressive capacity, we design four scenarios targeting specific affective quadrants of the Valence-Arousal space (Fig.~\ref{fig:VA_demo}).  To test baseline behavior and robustness, we added a fifth Neutral scenario and a sixth Ambiguous scenario designed to trigger the system's fallback mechanism.

\begin{itemize}
    \item S1: Aggression (Quadrant I). To test a high-arousal, negative reaction, the user faces the robot and makes a continuous punching motion in its direction.
    \item S2: Celebration (Quadrant II). To test high-arousal, positive expression, the user smiles broadly and raises both fists in a clear celebratory gesture.
    \item S3: Calm Greeting (Quadrant III). To test low-arousal, positive expression, the user gives a slow, gentle wave accompanied by a slight smile.
    \item S4: Disappointment (Quadrant IV).   To test a low-arousal, strong negative expression, the user, places their head in their hands and hunches their back forward in a gesture of distress.
    \item S5: Neutral. To establish a non-emotional baseline, the user, with a neutral expression, points towards a specific object in the environment.
    \item S6: Ambiguous. To validate the fallback mechanism, the user performs a gesture designed for ambiguity: a flat hand is raised and moved side-to-side or doing unintentional motions.
\end{itemize}

These six scenarios provide a comprehensive testbed for our evaluation. The four affective scenarios (S1-S4) are designed to specifically assess the performance of the V-A conditioning, which will be compared against a baseline system operating without affective reasoning in late sections. The two diagnostic scenarios (S5-S6) are used to validate the system's core functional capabilities and safety mechanisms, respectively, some cased are shown in Fig.~\ref{fig:HRI_Scenario}.

\subsection{Motion Generation Capabilities}
\begin{figure}
    \centering
    \includegraphics[trim = 0mm 91mm 200mm 0mm, clip, width=\linewidth]{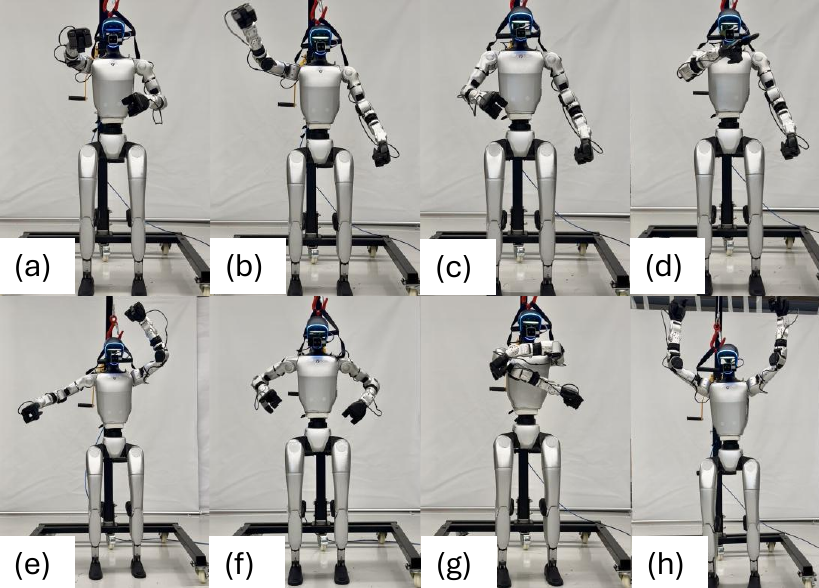}
    \caption{Diverse repertoire of social gestures generated and executed by HIAER. The figure showcases a selection of expressive motions produced by our text-to-motion pipeline and executed on physical robot in real time. The generated gestures represent a wide range of social signals, including (a) a defensive guard stance, (b) a greeting wave, (c) a handshake gesture, (d) a directional gesture to point, (e) an excited cheer, (f) a hands-on-hips stance, (g) a cross-armed pose, and (h) a two-armed celebration}
    \label{fig:motions}
\end{figure}

Before evaluating the complete HIAER framework in interactive scenarios, we first showcase the expressive range of our text-to-motion pipeline, which has been validated through our motion execution module with $\pi_p$ and $\pi_w$ on the physical robot. Our system is capable of generating a wide variety of socially relevant gestures based on textual prompts. Fig.\ref{fig:motions}  illustrates a selection of these generated motions, including emotional responses to greeting, defensive, and various emotional non-verbal gestures. 

\subsection{Socially-Aware Response Evaluation}
In this section, we present a quantitative and qualitative  evaluation of the HIAER framework, structured around three central research questions. Our goal is to validate HIAER's core contributions against a baseline system. This baseline also utilizes a VLM to select a motion $\langle m\rangle$ in an end-to-end fashion, but intentionally omits our structured, ICL-based reasoning module detailed in Section \ref{subsec:intention_aware_module}. This allows us to isolate and evaluate the specific benefits of our hierarchical inference approach.

\paragraph{Evaluation Metrics and Procedure}
To answer Q1, we conducted an evaluation of the intermediate outputs from the Intention-Aware Module $\pi_i$. The evaluation was performed across a total of $90$ trials, $15$ trails for each scenario (S1-S6). Three expert raters were invited to score the outputs. Before the main evaluation, a calibration session was held to ensure a consensus on scoring criteria. For each trial, the three raters were presented with the input video alongside the structured output from the $\pi_i$ module (including inferred intent $i$, the V-A pair $(\mathcal{V},\mathcal{A})$, and the selected motion primitive $\langle m\rangle$). They then provided scores for the following metrics using a 5-point Likert scale (1 = Very Inappropriate, 3 = Suitable, 5 = Very Appropriate):

\begin{itemize}
    \item \textbf{Inference Accuracy $I_\text{{acc}}$}: An objective metric measuring the percentage of trials where the module's inferred user intent $i$ correctly matched the pre-defined ground truth of the scenario.
    \item \textbf{Motion Selection $S_\text{{select}}$}: A subjective rating of the module's internal policy. Raters evaluated if the chosen motion primitive ⟨m⟩ was a logical and socially coherent response, conditioned on the module's own inferred intent and Valence-Arousal state $(\mathcal{V},\mathcal{A})$.
    \item \textbf{Affective Estimation $S_\text{{affect}}$}: A subjective rating of the overall interaction quality. Raters assessed the final appropriateness of the robot's motion within the interaction scenario.  
\end{itemize}

Our HIAER framework was evaluated on all three metrics. The baseline, which does not expose its internal reasoning state, could not be evaluated on $S_\text{{select}}$. 

\begin{table}[t]
\centering
\caption{Performance metrics of HIAER and baseline across scenarios. $I_{acc}$ is the objective intent inference accuracy. All other metrics are subjective 5-point Likert scale ratings, where higher is better. $V_{avg}$ and $A_{avg}$ are the averaged Valence and Arousal values inferred by HIAER.}
\label{tab:metrics+baseline}
\setlength{\tabcolsep}{6pt}
\renewcommand{\arraystretch}{1.15}
\begin{tabular}{c c c c c c !{\vrule width 0.6pt} c}
\toprule
\multirow{2}{*}{\textbf{Scenarios}}
& \multicolumn{5}{c}{\textbf{HIAER}} 
& \multicolumn{1}{c}{\textbf{Baseline}} \\
\cmidrule(lr){2-6} \cmidrule(lr){7-7}

& $I_{\text{acc}}$ 
& $V_{\text{avg}}$ 
& $A_{\text{avg}}$ 
& $S_{\text{select}}$ 
& $S_{\text{affect}}$ 
& $S_{\text{affect}}$ \\ 
\midrule
\centering{S1} 
   & 80.0\% & -0.46 & 0.64 & 5   & 4.42 & 4.16 \\
S2 & 93.3\% &  0.53 & 0.49 & 4.64 & 4.38 & 4.2 \\
S3 & 93.3\% &  0.36 & 0.29 & 4.2   & 4.87 & 4.82 \\
S4 & 93.3\% & -0.32 & 0.47 & 3.27   & 3.53 & 3.13 \\
S5 & 86.7\% &  0.03 & 0.25 & 4.87   & 4.4 & 3.71 \\
S6 & 80.0\% &  0.11 & 0.29 & 4.69   & 3.89 & 1.76 \\
\bottomrule
\end{tabular}
% \caption*{\textit{Note.}  $I_{acc}$ is the objective intent inference accuracy. All other metrics are subjective 5-point Likert scale ratings, where higher is better. $V_{avg}$ and $A_{avg}$ are the averaged Valence and Arousal values inferred by HIAER.}
\end{table}

\paragraph{Quantitative Results and Analysis} 
Table~\ref{tab:metrics+baseline}, demonstrate the effectiveness of HIAER and reveal its key advantages over the baseline, particularly in navigating social ambiguity. Overall, HIAER performs robustly in scenarios with clear affective cues. Within all scenarios, the $I_acc$ are above $80\%$. This accurate upstream inference directly translated to high downstream scores, with $S_\text{select}$ and affective appropriateness  $S_\text{affect}$ closely matching or exceeding the baseline's strong performance in these straightforward cases.

More critically, HIAER's hierarchical design provides crucial robustness when intent is ambiguous. In S1, S5 and S6, while the $I_acc$ is was understandably low, the system 's V-A estimation remained highly effective, correctly capturing the negative or neutral tone of the scene. This robust V-A conditioning acted as a ``social safety net,'' guiding the motion selection policy towards safe, neutral gestures, which were still rated as logical.

\begin{table}
\centering
\caption{Latency of each module in the real-Time pipeline}
\label{tab:latency_breakdown}
\begin{tabular}{ll}
\toprule
\textbf{Module} & \textbf{Latency}  \\
\midrule
Video Stream          & $20$ Hz\\
$\pi_i$  & $2.392$ (avg)  \\
$\pi_p$ & $0.087$ (avg per $8$ frames) \\
$\pi_w$        & $50$ Hz \\
\bottomrule
\end{tabular}
\end{table}

\paragraph{Qualitative Case Studies} To qualitatively illustrate HIAER’s behavior in real interactions, we present a case-by-case analysis of representative scenarios from Fig.~\ref{fig:HRI_Scenario}. This analysis demonstrates how the framework's internal reasoning, shown in the output boxes, directly leads to the socially intelligent behaviors quantified in Table \ref{tab:metrics+baseline}.

We first examine scenarios, the high-Arousal scenarios with Contrasting Valence cases, characterized by high emotional arousal but opposing valence `Aggressive' (S1) and `Celebration' (S2). In both cases, HIAER's intention-aware module correctly infers high arousal (A=$0.7$ in S1, A=$0.8$ in S2) with high confidence. Crucially, it distinguishes the negative valence (V=$-0.6$) of the aggressive posture from the positive valence (V=$0.6$) of the celebratory one. This distinction conditions the selection of appropriate high-energy motions: a defensive ``punch'' stance for S1 and an enthusiastic ``beat gesture'' for S2. Notably, in S2, the robot matches the user's positive affect rather than literally mimicking their pose, demonstrating a focus on conveying emotion over form.

In terms of low-Arousal scenarios, HIAER's capability extends to more subtle, low-energy social signals, as seen in scenarios S3, S4, and S5. In `Calm Greeting' (S3), its low-arousal inference (A=$0.3$) prompts a calm, reciprocal wave, achieving the highest subjective rating ($S_{\text{affect}}$ =$4.87$). This highlights that actions are chosen for social appropriateness, with the system showing response diversity by sometimes offering a handshake for the same intent. This contrasts with S4, where mirroring the ``cross arms'' gesture is a deliberate, strategic choice, used only because the inferred `Disappointing' context makes it an effective non-verbal cue of alignment. Similarly, in the `Neutral' scenario (S5), the framework correctly selects inaction (``stand still''), avoiding an unprompted and potentially awkward gesture.

% Although mirroring can be appropriate, in our system it is a deliberate and socially conditioned strategy rather than imitation. In S4, HIAER mirrors only when the inferred context is Disappointing, using the posture as a non-verbal cue of alignment. The framework first infers intent and estimates valence and arousal, and then conditions motion selection on the resulting social context rather than copying the raw pose. In S2, for example, the robot does not reproduce the user’s raised-arm posture but instead generates a beat gesture that conveys the same high-arousal and positive affect. For common intents such as S3, the system also demonstrates response diversity, sometimes offering a wave and in other instances a handshake, which highlights that actions are chosen for social appropriateness rather than pose similarity.

We also demonstrate an Handling Ambiguity with Confidence-Aware Selection. The framework's robustness is most evident under uncertainty. In S6, where the human's posture is unclear, the VLM outputs an `Ambiguous' intent with a very low confidence score. Recognizing its own uncertainty, HIAER defaults to the safest response: ``stand still.'' This cautious strategy, which prevents potentially inappropriate motions, is a key factor behind its significant performance advantage over the baseline in this scenario ($S_{\text{affect}}$ of $3.89$ vs. $1.76$).

\subsection{System performance analysis}
To answer Q3, we evaluate the system's practical viability by measuring its real-time system latency and assessing its robustness against common disturbances. Besides, we also validate how multi-modal input affects the accuracy of our intention awareness module.

\paragraph{System Latency} Our HIAER framework is designed as an asynchronous pipeline to achieve semi-real-time performance. A breakdown of the latency for each core module is presented in Table~\ref{tab:latency_breakdown}. As the data shows, the primary computational bottleneck is the VLM-based intention inference module $\pi_i$. Across 100 trials on our workstation, its latency averaged $2.392$ s (median: $2.25$ s; range: $1.72$ s–$2.83$ s). This latency is the main factor determining the robot's reaction time to new social cues. To manage this and prevent stale responses, we implement a $3$-second timeout mechanism: if an inference cycle exceeds this threshold, the next cycle begins with the most recent visual input.

\paragraph{Occlusion Disturbance} We evaluated the system's robustness to partial occlusions, a common challenge in dynamic environments. The evaluation involved three targeted conditions: facial, body, and limb occlusions. We observed that even when key visual features were obscured in instantaneous frames, the system's use of a temporal history of images allowed it to successfully infer intent from the remaining contextual and motion cues. Across these challenging occlusion conditions, the framework maintained a high success rate of approximately $80$\% in intent recognition.

\paragraph{Multi-person Disturbance} We also conducted exploratory tests in multi-person scenarios. Our system demonstrated the capability to detect and describe the presence of multiple individuals within the scene. However, we identified a current limitation in its intention-aware reasoning: it does not disambiguate between multiple, concurrent intentions. The system tends to focus on the most visually salient person or action and infers a single intent for the entire interaction.

\begin{table}
\centering
\caption{Ablation study on input modality.}
\begin{tabular}{c c}
\toprule
\textbf{Input Modality} & \textbf{Interaction Context Accuracy} \\
\midrule
Prompt only & 0.20 \\
Image only  & 0.77 \\
Combined    & 0.87 \\
\bottomrule
\end{tabular}
\label{tab:modality_ablation}
\end{table}

\paragraph{Multi-modal Input}
To evaluate the contribution of each input modality to human intention inference, we conduct an ablation study using ten challenging test cases involving ambiguous or low-confidence gestures—such as partial occlusion, subtle upper-body motion, or low-light conditions. We compare three input configurations: vision-only, language-only, and combined vision-language input. As summarized in Table~\ref{tab:modality_ablation}, the combined modality achieves substantially higher classification accuracy than either modality alone, underscoring the complementary strengths of visual and linguistic information for robust social context reasoning. This also supports the integration of fallback mechanisms when relying on single-modality input under uncertainty.

% \paragraph{history info}
% \hl{need extra experiment}

% The incorporation of historical context is critical for achieving robust and accurate human intention recognition. Human intent is often not conveyed in a single, isolated moment but unfolds dynamically over time. A static snapshot of a user's posture or expression can be inherently ambiguous, leading to incorrect inferences. By analyzing a sequence of observations, the system can disambiguate current actions by understanding the preceding events that provide crucial context. For instance, consider a scenario where a user fails to lift a heavy object (t1), sighs in frustration (t2), and then turns to look at the robot (t3). A system analyzing only the final frame (t3) might ambiguously infer a simple desire to initiate conversation. However, by integrating the historical context from t1 and t2, the intent to "seek help" becomes unequivocally clear. This temporal reasoning allows HIAER to move beyond simple reactive behaviors to a more profound, context-aware understanding of the user's needs.

% \hl{Need to do experiment}

% \begin{itemize}
%     \item Ambigous test
%     \item obstacles
%     \item multi-person
% \end{itemize}

\section{Conclusion and Future Study}
\label{sec:conclusion}

In this paper, we presented \textsc{HIAER}, a hierarchical framework that bridges high-level social reasoning with real-time motion generation to enable expressive, socially-aware responses on a physical humanoid robot. Our key finding is that by conditioning gestures on both inferred intention and V-A estimates, the robot achieves significantly improved social appropriateness and robustness. This was most evident in ambiguous scenarios where our approach produced safe, affectively coherent gestures while a non-hierarchical baseline failed, underscoring the critical role of affective context in navigating complex human-robot interactions. Besides, our physical experiment shows, they are robustness through disturbance of dynamics challenges.

We acknowledge our limitation on only upper-body expressive motions and lacking subjective impact on users. Our future research will proceed along two primary avenues. First, we will expand the robot's expressive repertoire to the whole body, incorporating posture shifts and subtle locomotion to convey a wider range of non-verbal cues. Second, we will conduct rigorous, formal user studies to evaluate the downstream impact of our framework on human perception, trust, and engagement in collaborative tasks.

\bibliographystyle{IEEEtran}
\bibliography{reference}

\begin{thebibliography}{10}
\providecommand{\url}[1]{#1}
\csname url@samestyle\endcsname
\providecommand{\newblock}{\relax}
\providecommand{\bibinfo}[2]{#2}
\providecommand{\BIBentrySTDinterwordspacing}{\spaceskip=0pt\relax}
\providecommand{\BIBentryALTinterwordstretchfactor}{4}
\providecommand{\BIBentryALTinterwordspacing}{\spaceskip=\fontdimen2\font plus
\BIBentryALTinterwordstretchfactor\fontdimen3\font minus \fontdimen4\font\relax}
\providecommand{\BIBforeignlanguage}[2]{{%
\expandafter\ifx\csname l@#1\endcsname\relax
\typeout{** WARNING: IEEEtran.bst: No hyphenation pattern has been}%
\typeout{** loaded for the language `#1'. Using the pattern for}%
\typeout{** the default language instead.}%
\else
\language=\csname l@#1\endcsname
\fi
#2}}
\providecommand{\BIBdecl}{\relax}
\BIBdecl

\bibitem{kendon2004gesture}
A.~Kendon, \emph{Gesture: Visible action as utterance}.\hskip 1em plus 0.5em minus 0.4em\relax Cambridge University Press, 2004.

\bibitem{cavallo2016decoding}
A.~Cavallo, A.~Koul \emph{et~al.}, ``Decoding intentions from movement kinematics,'' \emph{Scientific Reports}, vol.~6, no.~1, p. 37036, 2016.

\bibitem{2023_nonverbal_hri}
J.~Urakami and K.~Seaborn, ``Nonverbal cues in human--robot interaction: A communication studies perspective,'' \emph{ACM Transactions on Human-Robot Interaction}, vol.~12, no.~2, pp. 1--21, 2023.

\bibitem{2013_modeling_evaluation_humanoid_HRI}
C.-M. Huang and B.~Mutlu, ``Modeling and evaluating narrative gestures for humanlike robots.'' in \emph{Robotics: Science and Systems}, vol.~2, 2013.

\bibitem{2024_humanoid_LLM_HRI_understandingLLMpowered}
C.~Y. Kim, C.~P. Lee, and B.~Mutlu, ``Understanding large-language model ({LLM})-powered human-robot interaction,'' in \emph{ACM/IEEE International Conference on Human-robot Interaction}, 2024, pp. 371--380.

\bibitem{2024_Bao_drl_bipedal_review}
L.~Bao, J.~Humphreys \emph{et~al.}, ``Deep reinforcement learning for bipedal locomotion: A brief survey,'' \emph{arXiv preprint arXiv:2404.17070}, 2024.

\bibitem{2024_speratebody_expressive_allmotion}
X.~Cheng, Y.~Ji \emph{et~al.}, ``Expressive whole-body control for humanoid robots,'' in \emph{Proceedings of Robotics: Science and Systems}, 2024.

\bibitem{he2025asap}
T.~He, J.~Gao \emph{et~al.}, ``{ASAP}: Aligning simulation and real-world physics for learning agile humanoid whole-body skills,'' in \emph{Proceedings of Robotics: Science and Systems}, 2025.

\bibitem{2022_incontextlearning_rethinkingroleofdemonstration}
S.~Min, X.~Lyu \emph{et~al.}, ``Rethinking the role of demonstrations: What makes in-context learning work?'' in \emph{EMNLP}, 2022.

\bibitem{2024_ICL_EMOTION_humanoid_VLM_demonstration}
P.~Huang, Y.~Hu \emph{et~al.}, ``{EMOTION}: Expressive motion sequence generation for humanoid robots with in-context learning,'' \emph{IEEE Robotics and Automation Letters}, vol.~10, no.~8, pp. 7699--7706, 2025.

\bibitem{vamodel1}
S.~Sadeghi, Z.~Gu \emph{et~al.}, ``Direct perception of affective valence from vision,'' \emph{Nature Communications}, vol.~15, no.~1, p. 10735, 2024.

\bibitem{vamodel2}
V.~Ahire, K.~Shah \emph{et~al.}, ``{MAVEN}: Multi-modal attention for valence-arousal emotion network,'' in \emph{Proceedings of the Computer Vision and Pattern Recognition Conference Workshops}, June 2025, pp. 5789--5799.

\bibitem{2025_MDM_DART}
K.~Zhao, G.~Li, and S.~Tang, ``{DartControl}: A diffusion-based autoregressive motion model for real-time text-driven motion control,'' in \emph{The Thirteenth International Conference on Learning Representations}, 2025.

\bibitem{CoT}
J.~Wei, X.~Wang \emph{et~al.}, ``Chain-of-thought prompting elicits reasoning in large language models,'' \emph{Advances in Neural Information Processing Systems}, vol.~35, pp. 24\,824--24\,837, 2022.

\bibitem{2020_rule_based_Cassie_manual_interface}
Z.~Li, C.~Cummings, and K.~Sreenath, ``{Animated Cassie}: A dynamic relatable robotic character,'' in \emph{IEEE International Conference on Intelligent Robots and Systems}, 2020, pp. 3739--3746.

\bibitem{2013_HRI_rulebasedmethod_behaviorgeneneration}
A.~Aly and A.~Tapus, ``A model for synthesizing a combined verbal and nonverbal behavior based on personality traits in human-robot interaction,'' in \emph{IEEE International Conference on Human-Robot Interaction}, 2013, pp. 325--332.

\bibitem{2019_HRI_rulebased_Bodystorming}
D.~Porfirio, E.~Fisher \emph{et~al.}, ``Bodystorming human-robot interactions,'' in \emph{Proceedings of the 32nd Annual ACM Symposium on User Interface Software and Technology}, 2019, p. 479–491.

\bibitem{template1}
S.~Qian, Z.~Tu \emph{et~al.}, ``Speech drives templates: Co-speech gesture synthesis with learned templates,'' in \emph{IEEE/CVF International Conference on Computer Vision}, 2021, pp. 11\,057--11\,066.

\bibitem{VLM_intention1}
F.~Munir, S.~Azam \emph{et~al.}, ``Pedestrian vision language model for intentions prediction,'' \emph{IEEE Open Journal of Intelligent Transportation Systems}, vol.~6, pp. 393--406, 2025.

\bibitem{VLM_intention2}
C.~Cao, L.~Hu \emph{et~al.}, ``Vision and intention boost large language model in long-term action anticipation,'' \emph{arXiv preprint arXiv:2505.01713}, 2025.

\bibitem{2024_LLM_Prompt2Walk}
Y.-J. Wang, B.~Zhang \emph{et~al.}, ``Prompt a robot to walk with large language models,'' in \emph{IEEE Conference on Decision and Control}, 2024.

\bibitem{2024_LLM_humanoidrobot_ComprehensiveLocomotionControl}
S.~Sun, C.~Li \emph{et~al.}, ``Leveraging large language models for comprehensive locomotion control in humanoid robots design,'' \emph{Biomimetic Intelligence and Robotics}, vol.~4, no.~4, p. 100187, 2024.

\bibitem{2018_T2A_GANs}
H.~Ahn, T.~Ha \emph{et~al.}, ``{Text2Action}: Generative adversarial synthesis from language to action,'' in \emph{IEEE International Conference on Robotics and Automation}, 2018, pp. 5915--5920.

\bibitem{2020_action2motion_VAEs}
C.~Guo, X.~Zuo \emph{et~al.}, ``{Action2Motion}: Conditioned generation of {3D} human motions,'' in \emph{Proceedings of the 28th ACM International Conference on Multimedia}, 2020, pp. 2021--2029.

\bibitem{2020_diffusion_model}
J.~Ho, A.~Jain, and P.~Abbeel, ``Denoising diffusion probabilistic models,'' \emph{Advances in neural information processing systems}, vol.~33, pp. 6840--6851, 2020.

\bibitem{2023_tevet_MDM}
G.~Tevet, S.~Raab \emph{et~al.}, ``Human motion diffusion model,'' in \emph{The Eleventh International Conference on Learning Representations}, 2023.

\bibitem{2019_Naureen_AMASS}
N.~Mahmood, N.~Ghorbani \emph{et~al.}, ``{AMASS}: Archive of motion capture as surface shapes,'' in \emph{IEEE International Conference on Computer Vision}, 2019, pp. 5441--5450.

\bibitem{2022_Dataset_HumanML3D_Guo_CVPR}
C.~Guo, S.~Zou \emph{et~al.}, ``Generating diverse and natural 3d human motions from text,'' in \emph{Proceedings of the IEEE Conference on Computer Vision and Pattern Recognition}, June 2022, pp. 5152--5161.

\bibitem{2024_GenerativeExpressiveBehavior_LLM}
K.~Mahadevan, J.~Chien \emph{et~al.}, ``Generative expressive robot behaviors using large language models,'' in \emph{Proceedings of the ACM/IEEE International Conference on Human-Robot Interaction}, 2024, p. 482–491.

\bibitem{Qwen2.5_VL}
S.~Bai, K.~Chen \emph{et~al.}, ``{Qwen2.5-VL} technical report,'' \emph{arXiv preprint arXiv:2502.13923}, 2025.

\bibitem{BABEL}
A.~R. Punnakkal, A.~Chandrasekaran \emph{et~al.}, ``{BABEL}: Bodies, action and behavior with english labels,'' in \emph{Proceedings of the IEEE/CVF Conference on Computer Vision and Pattern Recognition}, 2021, pp. 722--731.

\bibitem{SMPL}
M.~Loper, N.~Mahmood \emph{et~al.}, ``{SMPL}: A skinned multi-person linear model,'' \emph{ACM Trans. Graphics (Proc. SIGGRAPH Asia)}, vol.~34, no.~6, pp. 248:1--248:16, Oct. 2015.

\bibitem{2019_Cont6d}
Y.~Zhou, C.~Barnes \emph{et~al.}, ``On the continuity of rotation representations in neural networks,'' in \emph{Proceedings of the IEEE conference on computer vision and pattern recognition}, 2019, pp. 5745--5753.

\bibitem{unitree_g1}
{Unitree Robotics}, ``Unitree {G1} humanoid robot,'' \url{https://www.unitree.com/g1}, 2024, accessed: 2025-09-13.

\bibitem{he2024learning}
T.~He, Z.~Luo \emph{et~al.}, ``Learning human-to-humanoid real-time whole-body teleoperation,'' in \emph{IEEE/RSJ International Conference on Intelligent Robots and Systems}, 2024.

\bibitem{isaaclab}
M.~Mittal, C.~Yu \emph{et~al.}, ``Orbit: A unified simulation framework for interactive robot learning environments,'' \emph{IEEE Robotics and Automation Letters}, vol.~8, no.~6, pp. 3740--3747, 2023.

\bibitem{daily_emotion}
A.~S. Cowen and D.~Keltner, ``Self-report captures 27 distinct categories of emotion bridged by continuous gradients,'' \emph{Proceedings of the national academy of sciences}, vol. 114, no.~38, pp. E7900--E7909, 2017.

\end{thebibliography}

\end{document}